\documentclass[preprint,12pt]{elsarticle}
\journal{Computers, Environment and Urban Systems}
\usepackage{url}
\usepackage{natbib}
\usepackage{hyperref}
\usepackage{graphicx}
\usepackage[export]{adjustbox}
\usepackage{float}
\usepackage{tabularx}
\usepackage{booktabs}
\usepackage{caption}
\usepackage{fancyhdr}
\usepackage{pdflscape}
\usepackage{amsmath}
\usepackage{longtable}
\usepackage{geometry}
\usepackage{fancyhdr}
\usepackage{setspace}
\makeatletter
\renewcommand\subsection{\@startsection{subsection}{2}{\z@}%
           {12\p@ \@plus 6\p@ \@minus 3\p@}%
           {3\p@ \@plus 6\p@ \@minus 3\p@}%
           {\normalfont\normalsize\bfseries}}   
\renewcommand\subsubsection{\@startsection{subsubsection}{3}{\z@}%
           {12\p@ \@plus 6\p@ \@minus 3\p@}%
           {3\p@ \@plus 6\p@ \@minus 3\p@}%
           {\normalfont\normalsize\bfseries}}   
\makeatother

\usepackage{enumitem}
\usepackage{pgfplots}
\pgfplotsset{compat=1.18}
\usepackage{adjustbox}
\usepackage{multirow} 
\usepackage{array}
\usepackage[export]{adjustbox}
\usepackage{xcolor}
\usepackage{xurl}
\usepackage{makecell}
\usepackage{threeparttable} 

\begin{document}
\fancyhf{} 
\fancyfoot[R]{\thepage} 
\pagestyle{fancy}
\renewcommand{\headrulewidth}{0pt} 
\renewcommand{\footrulewidth}{0pt} 
\begin{frontmatter}

\title{\textbf{Towards Accurate and Efficient Waste Image Classification: A Hybrid Deep Learning and Machine Learning Approach}}
\author{Ngoc-Bao-Quang Nguyen}

\author{Tuan-Minh Do}

\author{Cong-Tam Phan}

\author{Thi-Thu-Hong Phan\corref{cor1}}
\ead{hongptt11@fe.edu.vn}

\address{Department of Artificial Intelligence, FPT University, Da Nang, 550000, Viet Nam}

\cortext[cor1]{Corresponding author}

\begin{abstract}
Automated image-based garbage classification is a critical component of global waste management; however, systematic benchmarks that integrate Machine Learning (ML), Deep Learning (DL), and efficient hybrid solutions remain underdeveloped. This study provides a comprehensive comparison of three paradigms: (1) machine learning algorithms using handcrafted features, (2) deep learning architectures, including ResNet variants and EfficientNetV2S, and (3) a hybrid approach that utilizes deep models for feature extraction combined with classical classifiers such as Support Vector Machine and Logistic Regression to identify the most effective strategy. Experiments on three public datasets—TrashNet, Garbage Classification, and a refined Household Garbage Dataset (with 43 corrected mislabels)—demonstrate that the hybrid method consistently outperforms the others, achieving up to 100\% accuracy on TrashNet and the refined Household set, and 99.87\% on Garbage Classification, thereby surpassing state-of-the-art benchmarks. Furthermore, feature selection reduces feature dimensionality by over 95\% without compromising accuracy, resulting in faster training and inference. This work establishes more reliable benchmarks for waste classification and introduces an efficient hybrid framework that achieves high accuracy while reducing inference cost, making it suitable for scalable deployment in resource-constrained environments.
\end{abstract}

\begin{keyword}
Waste classification, Deep learning, Machine learning, Feature extraction, Hybrid approach
\end{keyword}

\end{frontmatter}

\section{Introduction}
\label{sec1}

The exponential growth of household and industrial waste has become a pressing global concern, placing immense pressure on waste management systems and accelerating the depletion of landfill capacity. Waste sorting is widely recognized as the cornerstone of sustainable management, as it enables efficient recycling, resource recovery, and reduction of environmental burdens. However, manual sorting remains the predominant practice in many contexts, making the process costly, time-consuming, and inefficient.

To address this challenge, numerous studies have investigated the application of Artificial Intelligence (AI) for automated waste classification. Machine learning (ML) approaches that rely on handcrafted features and classical classifiers have shown promising results in certain cases \citep{fu2025}, while deep learning (DL) architectures, particularly Convolutional Neural Networks (CNNs), have achieved state-of-the-art performance in image-based tasks \citep{aghilan2020,huang2020, wulansari2022, wang2024}. More recently, hybrid approaches that integrate deep features with traditional classifiers have demonstrated superior performance \citep{celik, nahiduzzaman2025, li2025, dao2025}.

However, most of these studies address isolated aspects of the problem rather than providing an integrated benchmark across paradigms. Systematic comparisons that jointly evaluate ML, DL, and hybrid approaches remain largely unexplored in the waste classification literature. In particular, few studies have investigated how hybrid frameworks—especially those enhanced by feature selection—can balance predictive accuracy and computational efficiency. Furthermore, data quality issues in widely used public datasets (e.g., mislabeling in the Household Garbage dataset) have not been rigorously addressed, potentially compromising benchmark reliability.

To bridge these gaps, this study conducts a comprehensive and unified evaluation of three paradigms-classical ML using handcrafted features, DL models, and a hybrid DL–ML framework—for image-based waste classification. The proposed hybrid framework extracts deep representations from a customized ResNet50 backbone and integrates them with ML classifiers such as Support Vector Machine (SVM) and Logistic Regression (LR). In addition, feature selection is employed to enhance model compactness and inference efficiency, while dataset refinement through relabeling improves benchmark reliability. Collectively, these components form a more robust and computationally efficient soft computing pipeline for practical deployment.

The main contributions of this work are summarized as follows:

1. Comprehensive benchmarking: This work provides a more comprehensive benchmark across three public waste image datasets (TrashNet, Garbage Classification, and Household Garbage), jointly evaluating machine learning models based on handcrafted features, deep learning architectures, and hybrid deep–machine learning frameworks.

2. Dataset correction: A manual inspection and relabeling of 43 misclassified images in the Household Garbage dataset improves data integrity and ensures more reliable evaluation.

3. Hybrid framework: A hybrid design that combines deep features extracted from a customized ResNet50 with feature selection achieves over 95\% dimensionality reduction while maintaining near-perfect accuracy.

4. Efficiency and scalability analysis: Extensive experiments demonstrate that the proposed hybrid model attains state-of-the-art accuracy with substantially lower inference cost, making it suitable for resource-constrained or real-time intelligent systems.

The remainder of this paper is structured as follows. Section~\ref{sec:related} reviews related work on deep learning and hybrid frameworks for waste classification. Section~\ref{sec:methodology} describes the proposed approach, including architecture design and feature selection. Section~\ref{sec:results} reports the experimental setup and comparative evaluations. Section~\ref{sec:discussion} analyzes the results in terms of robustness and efficiency. Finally, Section~\ref{sec:conclusion} summarizes the findings and outlines future directions.

\section{Related work}
\label{sec:related}

In recent years, waste classification using image-based methods has gained considerable attention due to its potential in automating waste sorting and promoting sustainable waste management. Early efforts primarily relied on classical machine learning pipelines, typically following a two-step process: handcrafted feature extraction, most commonly based on local descriptors such as the Scale-Invariant Feature Transform (SIFT), followed by classification using models like Support Vector Machines (SVM) \citep{puspaningrum2020siftpca}. While such approaches showed initial promise, their performance was limited by the reliance on manually engineered features, which often struggled to generalize to diverse real-world waste items. Recent analyses, such as \citep{fu2025}, further confirmed these trade-offs: classical ML pipelines, including Support Vector Machine (SVM) and Decision Trees (DT), achieved only around 80–85\% accuracy on benchmark garbage datasets, significantly lower than CNN-based models exceeding 90\%. These results highlight the historical importance of traditional ML methods but also reveal their fundamental limitation: dependence on handcrafted features makes them less scalable and less robust when faced with the diversity and noise of real-world waste streams.

The advent of deep learning, particularly Convolutional Neural Networks (CNNs), marked a paradigm shift in image classification. Numerous studies have demonstrated 
the effectiveness of transfer learning, where pre-trained CNNs are fine-tuned to classify waste categories with high accuracy. Early supervised deep learning techniques also showed promise: for instance, \citep{aghilan2020} demonstrated the feasibility of CNN-based classification on garbage images, laying the foundation for later transfer learning works. Building on this, \citep{huang2020} implemented a model fusion approach using VGG19, DenseNet169, and NASNetLarge, achieving up to 96.5\% accuracy across datasets. In 2021, \citep{shi2021} addressed efficiency by proposing the Multilayer Hybrid Convolution Neural Network, a simplified architecture that achieved 92.6\% accuracy on the TrashNet dataset.

The following years saw the application of deep learning in more specialized domains. For example, \citep{wulansari2022} applied transfer learning to medical waste classification and reported 99.40\% accuracy. In 2023, several studies explored diverse hybrid and comparative strategies: \citep{wang2024} 
combined Error-Correcting Output Codes (ECOC) with Artificial Neural Networks (ANN), reaching 99.01\% accuracy and further enhancing CNN features using the Capuchin Search Algorithm (CapSA). \citep{lilhore2023} introduced a hybrid CNN-LSTM model that achieved 95.45\% accuracy for binary waste classification. \citep{mehedi2023} compared popular CNNs including VGG16 and MobileNetV2, showing that VGG16 performed better in their context. In the same year, \citep{sharma2023} applied deep CNNs for garbage classification, reinforcing the effectiveness of end-to-end pipelines in this domain. Nevertheless, while DL-based methods consistently outperform classical ML in accuracy, their dependence on large-scale pretraining, high computational cost, and limited inference efficiency remains a key challenge for real-time deployment 
and scalability in resource-constrained environments.

More recent works have focused on performance–efficiency trade-offs. \citep{kunwar2024} evaluated multiple transfer learning models and highlighted EfficientNetV2S for its strong balance of accuracy (96.41\%) and low carbon emissions. \citep{sayem2024} introduced a dual-stream architecture combining DenseNet201 and MaxViT for both classification and detection, though it reported a modest classification accuracy of 83.11\%, illustrating the trade-off between accuracy and scalability. \citep{celik} presented a sophisticated hybrid framework integrating EfficientNetB0 and InceptionV3 with a HyperColumn technique, achieving 99.40\% accuracy on TrashNet and 99.87\% on the Household Garbage dataset, thereby establishing a new benchmark. In parallel, \citep{nahiduzzaman2025} introduced a three-stage framework using parallel depth-wise separable CNNs (DP-CNN), showing that lightweight architectures can also achieve competitive performance with improved efficiency. Another recent development, \citep{li2025} proposed a ResNet50 model tailored for efficient waste classification. Their architecture leverages two key strategies: redundancy-weighted feature fusion and depth-separable convolutions. This design specifically aims to balance model complexity and classification performance. Evaluated on the TrashNet dataset, the refined model not only achieved a competitive accuracy of 94.13\% but also demonstrated a marked reduction in parameter count and improved inference efficiency.

These advancements highlight the promise of DL and hybrid strategies for waste classification. Nevertheless, systematic benchmarks that jointly consider accuracy, efficiency, and scalability are still lacking, motivating our evaluation in the following sections.

\section{Methodology}
\label{sec:methodology}
\subsection{Overview of methodology}

\begin{figure}[htbp]
        \begin{center}
            \rotatebox{0}{\includegraphics[scale = 0.37]{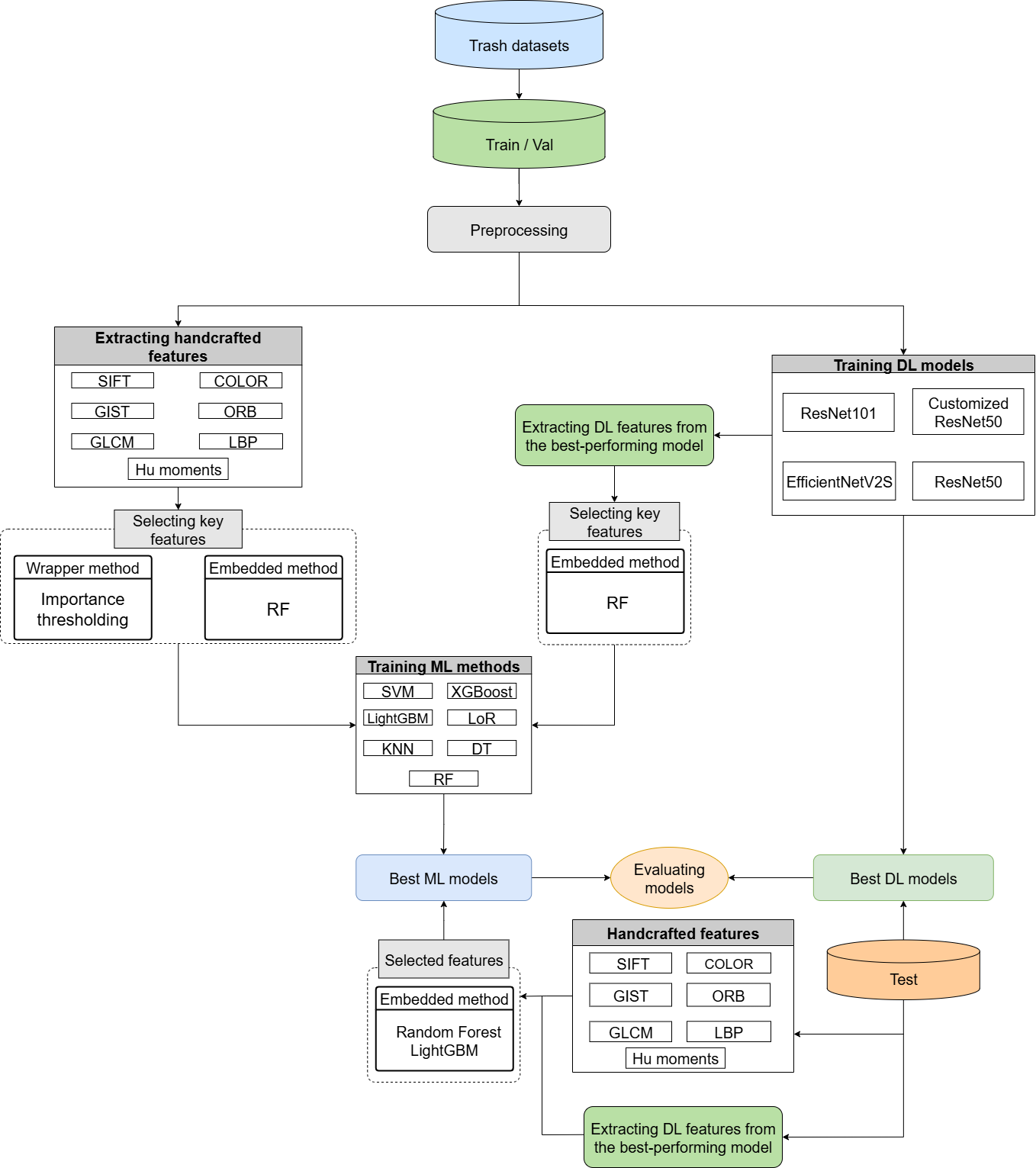}}
            \caption{Overview of proposed approach for garbage image classification}
            \label{fig:pipeline}
        \end{center}
    \end{figure}

Figure \ref{fig:pipeline} presents a comprehensive hybrid machine learning pipeline for waste classification. The pipeline is designed to explore and combine two distinct approaches, handcrafted features and deep learning features, to achieve optimal classification performance.

1. Approach based on handcrafted features

The pipeline begins with data pre-processing and splitting the dataset into training and validation sets. The first main branch of the pipeline focuses on traditional computer vision methods. Various handcrafted features are extracted from the images, including Scale-Invariant Feature Transform (SIFT), Color histograms, GIST descriptor, Oriented FAST and Rotated BRIEF (ORB), Gray-Level Co-occurrence Matrix (GLCM), and Local Binary Patterns (LBP). These features are then fed into a feature selection stage, utilizing both Wrapper and Embedded methods (such as importance thresholding and Random Forest). The selected features are subsequently used to train a range of classical machine learning models, including Support Vector Machine (SVM), Extreme Gradient Boosting (XGBoost), Logistic Regression (LoR), K-Nearest Neighbors (KNN), Decision Tree (DT), and Random Forest (RF). The best-performing models from this process are then evaluated on the test set.

2. Approach based on deep learning and a hybrid method

The second branch of the pipeline involves training several deep learning models, specifically ResNet101, EfficientNetV2S, and a customized ResNet50. These models are directly trained for the classification task, and their performance is evaluated to identify the best-performing DL models.

A key aspect of this pipeline is the hybrid approach, as illustrated by the "Extracting DL features" step. The best-performing model from the deep learning training phase is selected, and its learned features are extracted from a specific layer. This process provides powerful, high-level representations of the input images. Similar to the handcrafted approach, these deep features are then passed through a feature selection stage to refine the feature set. Finally, the selected deep features are used to train the same set of traditional machine learning models (SVM, XGBoost, etc.) to evaluate if this hybrid approach yields superior results by combining the strong feature learning capabilities of deep neural networks with the robust classification performance of classical ML algorithms. The following section will detail the specific techniques used in this study.

\subsection{Feature extraction algorithms}

In our study, we employed a two-stage approach for feature extraction to prepare the input data for machine learning models. The first stage, segmentation, focused on isolating the objects of interest from the background, while the second stage, handcrafted feature extraction, involved computing a variety of visual descriptors from the segmented images.

\subsubsection{Segmentation}
Classical computer vision techniques were used to segment the images and isolate the object of interest from the background. Specifically, the pipeline includes:

\begin{itemize}
    \item \textbf{GrabCut algorithm:} A semi-automatic segmentation technique based on graph cuts and Gaussian Mixture Models (GMM) was applied to separate the foreground from the background. A bounding rectangle was used to initialize the process, and the algorithm iteratively refined the segmentation \citep{grabcut}.
    
    \item \textbf{Thresholding and contour detection:} After the initial segmentation, the image was converted to grayscale and thresholded to obtain a binary mask. Contour detection was then applied to identify the largest connected component (assumed to be the object), and a bounding rectangle was computed to tightly crop it.
\end{itemize}

\subsubsection{Handcrafted feature extraction}
A variety of classical computer vision techniques were employed to extract relevant features from the segmented images. 
Each method was designed to capture different visual characteristics, including color, shape, texture, and keypoints. 
The extracted feature dimensions were as follows:
\begin{itemize}
 \item \textbf{Color features:}

Color basic:  To capture fundamental color properties, we calculated five statistical moments (mean, standard deviation, skewness, kurtosis, and entropy) for each of the three channels in the HSV color space. This process yields a 15-dimensional feature vector \citep{color_basic}.

Color histogram: This feature group captures the fine-grained color distribution of image pixels through 3D histograms computed in HSV and BGR color spaces. The concatenation of these two histograms results in a 1024-dimensional feature vector \citep{color_histogram}.

 \item \textbf{Shape features:}

Contour-based features: This feature set captures geometric properties of the object's primary contour, such as \textit{area, perimeter, aspect ratio, extent,} and \textit{solidity}. The analysis results in a compact 5-dimensional feature vector \citep{contour}.

Hu moments: Hu moments are calculated from the masked grayscale image to capture shape characteristics. A key advantage of this method is its invariance to translation, scale, and rotation, resulting in a 7-dimensional feature vector \citep{hu_moments}.

 \item \textbf{Texture features:}

\vspace{0.2cm}

Gray-Level Co-occurrence Matrix (GLCM): To capture the spatial relationships between pixel intensities, we calculated properties from the GLCM such as \textit{contrast, dissimilarity, homogeneity, energy,} and \textit{correlation} across four different angles. This process extracts a 20-dimensional feature vector \citep{glcm}.

\vspace{0.2cm}

Local Binary Pattern (LBP): The masked grayscale image to capture local texture. The operator functions by comparing each pixel to its neighbors, and a histogram of the resulting patterns forms a 10-dimensional feature vector \citep{lbp}.

\vspace{0.2cm}

 \item \textbf{Keypoint-based features:}
\vspace{0.2cm}

Oriented FAST and Rotated BRIEF (ORB): The algorithm was used on the masked grayscale image to detect keypoints and compute their descriptors. Each keypoint is described by a 32-byte binary vector \citep{orb}.

\vspace{0.2cm}

Scale-Invariant Feature Transform (SIFT): Keypoints and their corresponding descriptors were extracted from the masked grayscale image using the SIFT (Scale-Invariant Feature Transform) algorithm. Each keypoint is represented by a 128-dimensional feature vector \citep{sift}.
\vspace{0.2cm}

 \item \textbf{Global features:}

\vspace{0.2cm}

GIST descriptor: The image was first divided into a 4x4 grid. A set of Gabor filters with four orientations was then applied to each cell of the grid. The filter responses from all cells were concatenated to form a final 64-dimensional feature vector\citep{gist_descriptor}.
\end{itemize}

Finally, all the aforementioned feature sets were concatenated to form a single, comprehensive handcrafted feature vector for each image. This final vector combines color features (1039 dimensions), shape features (12 dimensions), texture features (30 dimensions), keypoint-based features (32 from ORB + 128 from SIFT), and global features (64 dimensions). 

In total, this process results in a high-dimensional feature vector of 1305 dimensions, which encapsulates a rich variety of visual information and is used as input for the classical machine learning classifiers.

\subsection{Machine learning methods}
   
After feature extraction, the obtained features are fed into a diverse set of classifiers, ranging from linear models to ensemble methods, in order to evaluate their effectiveness on high-dimensional image-derived data.

\begin{itemize}
    \item \textbf{Logistic Regression (LoR):} 
    A linear model for binary and multiclass classification that estimates class probabilities using the logistic (sigmoid) function. It serves as a strong and interpretable baseline for linearly separable data~\citep{logistic_regression}.

    \item \textbf{K-Nearest Neighbors (KNN):} 
    A simple yet effective non-parametric method that classifies a sample based on the majority class among its $K$ nearest neighbors in the feature space. It works well for problems with clearly separated clusters~\citep{knn}.
    
    \item \textbf{Support Vector Machine (SVM):} 
    A powerful classifier that constructs hyperplanes in a high-dimensional space to separate classes with maximum margin. It is particularly effective in high-dimensional settings, especially when the number of features exceeds the number of samples~\citep{svm}.

    \item \textbf{Decision Tree (DT):} 
    A tree-structured model that splits the data based on feature thresholds to create interpretable decision rules. While easy to understand and visualize, single decision trees often suffer from overfitting~\citep{decision_tree}.

    \item \textbf{Random Forest (RF):} 
    An ensemble method that builds multiple decision trees on bootstrap samples with feature randomness. It aggregates their predictions to improve generalization, reduce variance, and enhance robustness against overfitting and noise~\citep{random_forest}.

    \item \textbf{Extreme Gradient Boosting (XGBoost):} 
    A highly optimized and regularized gradient boosting technique that builds additive tree models in a sequential manner. It supports parallel computation and includes L1/L2 regularization to mitigate overfitting~\citep{xgboost}.
    
    \item \textbf{Light Gradient Boosting Machine (LightGBM):} 
    A fast, efficient implementation of gradient boosting that grows trees leaf-wise using histogram-based algorithms. It supports categorical features natively and is highly scalable for large datasets, making it well-suited for structured data derived from image descriptors~\citep{lightgbm}.

\end{itemize}

\subsection{Deep learning models}

Our model selection was guided by the objective of conducting a fair and representative comparison of CNN architectures for waste classification. Following insights from Kunwar et al.~\citep{kunwar2024}, who demonstrated the effectiveness of ResNet and EfficientNet variants, we selected three  deep learning models: the standard ResNet50 as a robust baseline, a customized ResNet50 with a task-specific classification head to leverage pre-trained features efficiently, and EfficientNetV2S, which emphasizes an optimal balance between accuracy and computational efficiency. This combination allows us to compare classical and customized ResNet designs alongside a modern, efficient architecture, providing a comprehensive assessment of CNN-based approaches for our task.

\begin{itemize}
   \item \textbf{Residual Networks}: 
    The ResNet family~\citep{resnet} introduces residual connections to mitigate the vanishing gradient problem and stabilize the training of deep architectures. 
    Both ResNet50 and ResNet101 are widely adopted convolutional neural networks that balance accuracy and computational cost at different depths. 
    While ResNet50 offers an efficient backbone for transfer learning with moderate complexity, ResNet101 provides a deeper variant with enhanced feature extraction capacity, suitable for capturing more fine-grained visual patterns. 
    Pre-trained on large-scale datasets such as ImageNet, these models deliver strong feature representations that can be effectively adapted for downstream tasks like waste classification. In this study, we experiment with both ResNet50 and ResNet101 architectures to evaluate their effectiveness for the waste classification task.

    \item \textbf{Customized ResNet50}: 
    Our customized model builds upon the pre-trained ResNet50 backbone. The original top classification layers are replaced with a GlobalAveragePooling2D layer followed by a dense output layer corresponding to the number of target classes. This design preserves the robust high-level features learned from ImageNet while creating a task-specific, lightweight classifier that can be efficiently fine-tuned for waste classification.

    To enhance model stability and improve learning on hard-to-classify samples, 
    the customized ResNet50 was trained using the \textit{Sparse Categorical Focal Loss}. 
    This loss extends the standard Cross-Entropy by introducing a modulating factor 
    that down-weights well-classified examples, enabling the model to focus on more challenging instances. 
    The formulation (\ref{eq:focal_loss_en}) proposed by ~\citep{lin2017focalloss} is defined as:

    \begin{equation}
    \label{eq:focal_loss_en}
    \mathcal{L}_{\text{FL}}(p_t) = - \alpha_t (1 - p_t)^\gamma \log(p_t)
    \end{equation}
    where $p_t$ is the predicted probability of the true class, $\gamma$ is the focusing parameter, 
    and $\alpha_t$ is the class-specific weight (default $\alpha = 0.25$). 
    This configuration enhances robustness against imbalance and improves performance 
    across minority waste categories without requiring additional sampling techniques.

    \item \textbf{EfficientNetV2S}: 
    EfficientNetV2S ~\citep{efficientnetv2} is a modern CNN architecture optimized for both accuracy and computational efficiency. Unlike ResNet, which primarily scales depth, EfficientNetV2 employs a compound scaling strategy that adjusts network depth, width, and input resolution simultaneously. The "S" (Small) variant was chosen to focus on architectures suitable for practical deployment, particularly in scenarios with limited computational resources, offering fast inference and scalable deployment while maintaining state-of-the-art performance.
\end{itemize}

\subsection{Hybrid approach}

The hybrid approach leverages the feature extraction capabilities of deep learning while combining them with the efficiency and flexibility of classical machine learning classifiers. In this framework, the top-performing  CNN model (either customized ResNet50 or EfficientNetV2S) is used solely as a feature extractor. For each input image, the output of the GlobalAveragePooling2D layer is extracted, producing a 2048-dimensional feature vector that encodes high-level semantic information.

These feature vectors are then processed through a feature selection pipeline to identify the most informative components, reducing redundancy and improving downstream learning efficiency. The selected features are fed into classical ML classifiers, such as SVM, LoR, or RF. This separation allows the ML models to learn specialized decision boundaries without retraining the full deep network, resulting in significantly reduced per-sample inference time.

Overall, the hybrid pipeline combines the representational power of deep networks with the lightweight, adaptable nature of traditional ML methods. It is particularly advantageous in scenarios requiring low-latency inference or deployment on resource-constrained devices, while maintaining classification accuracy comparable to  CNN models.

\section{Experimental results}
\label{sec:results}
\subsection{Datasets}
To ensure a comprehensive and robust evaluation, our study utilizes three distinct, publicly available datasets. Each dataset differs in terms of class diversity, image composition, and scale, allowing us to rigorously assess the generalizability of the proposed models.

\vspace{0.2cm}
\textbf{Dataset A - Garbage Classification} 

This dataset \citep{garbage_classification_dataset} consists of 19,762 images categorized into 10 waste classes. It is widely used for classification and object detection tasks, and is particularly relevant for recycling and waste management applications.

\begin{figure}[H]
    \centering
    \includegraphics[width=1\textwidth]{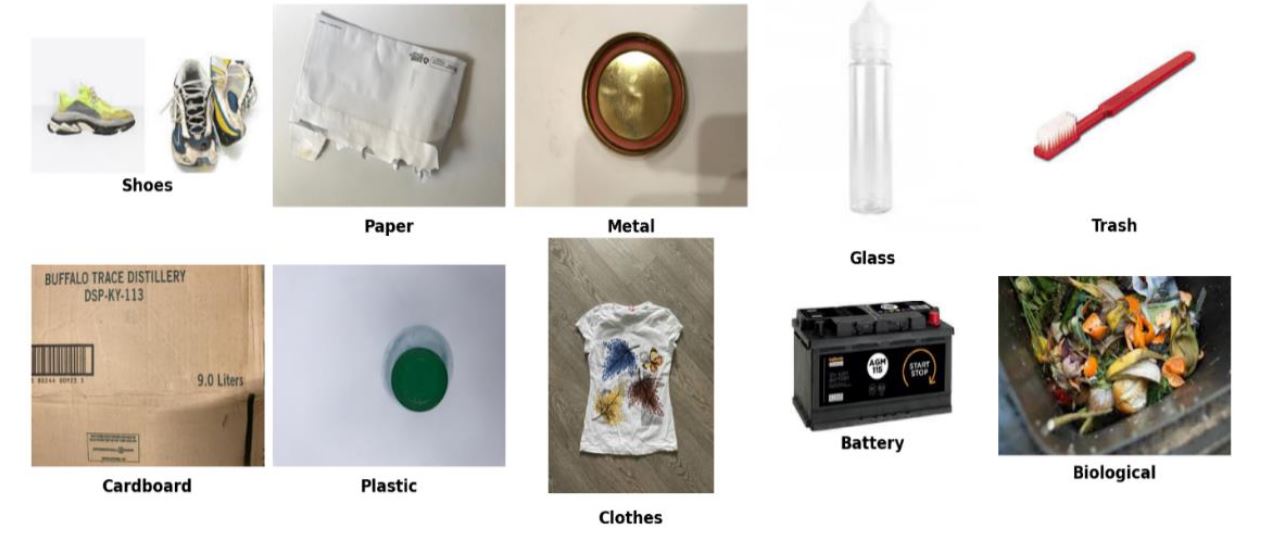}
    \caption{Sample images from the Garbage Classification dataset}
    \label{fig:primary_data}
\end{figure}

\textbf{Dataset B - Household Garbage}

This dataset \citep{household_garbage_classification} contains 15,150 images categorized into 12 classes of household garbage. It includes a broader range of categories, such as multiple types of glass, making it suitable for more fine-grained classification tasks.

\begin{figure}[H]
    \centering
    \includegraphics[width=0.8\textwidth]{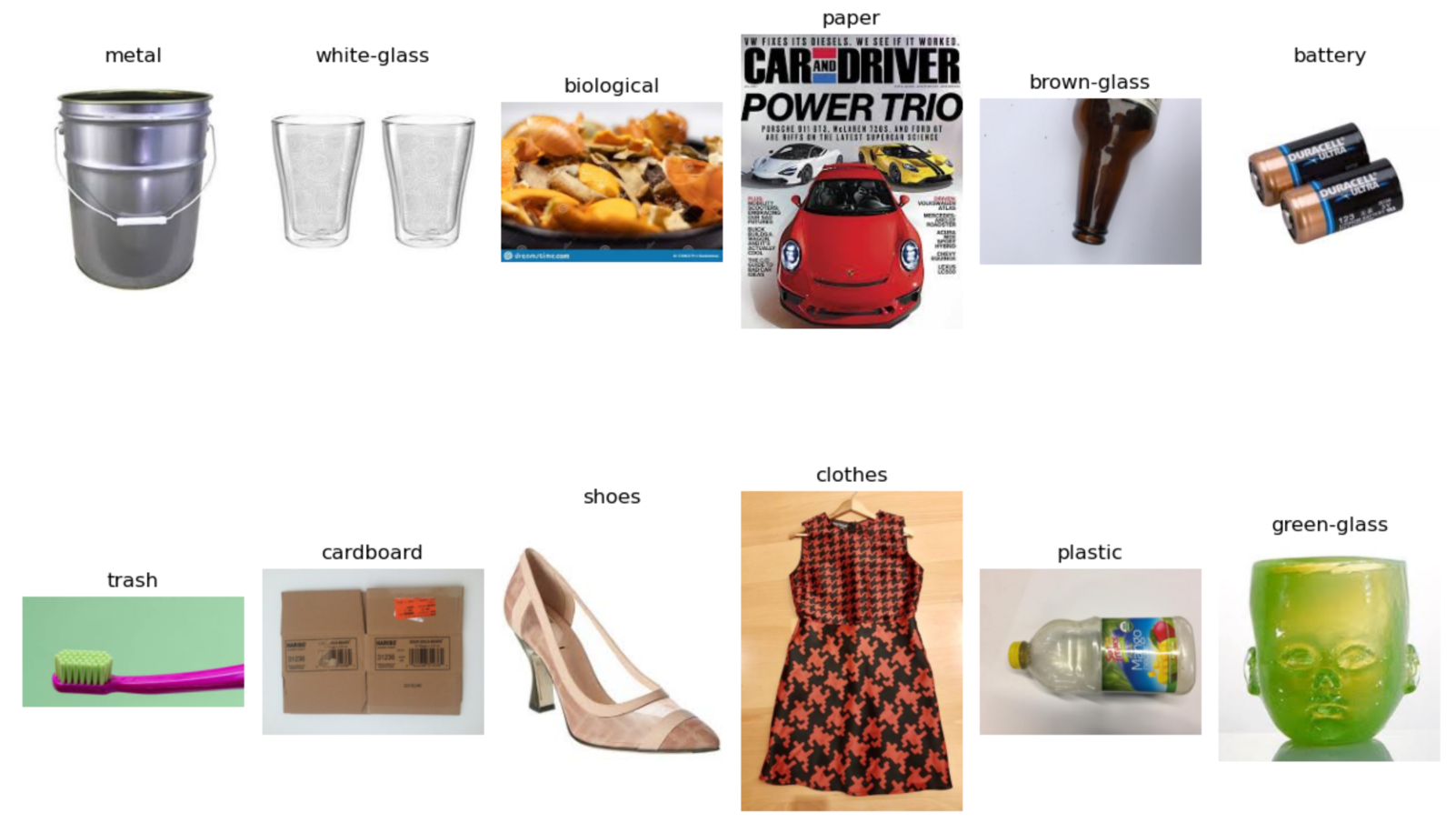}
    \caption{Sample images from the Household Garbage dataset}
    \label{fig:data_2}
\end{figure}

\textbf{Dataset C - TrashNet:}  

The TrashNet dataset \citep{trashnet} is a widely adopted public benchmark for waste classification. It comprises 2,527 images distributed across 6 material categories: \textit{glass, paper, cardboard, plastic, metal}, and \textit{trash}. Despite its smaller size, it remains an important reference dataset for baseline evaluation.

\begin{figure}[H]
    \centering
    \includegraphics[width=0.7\textwidth]{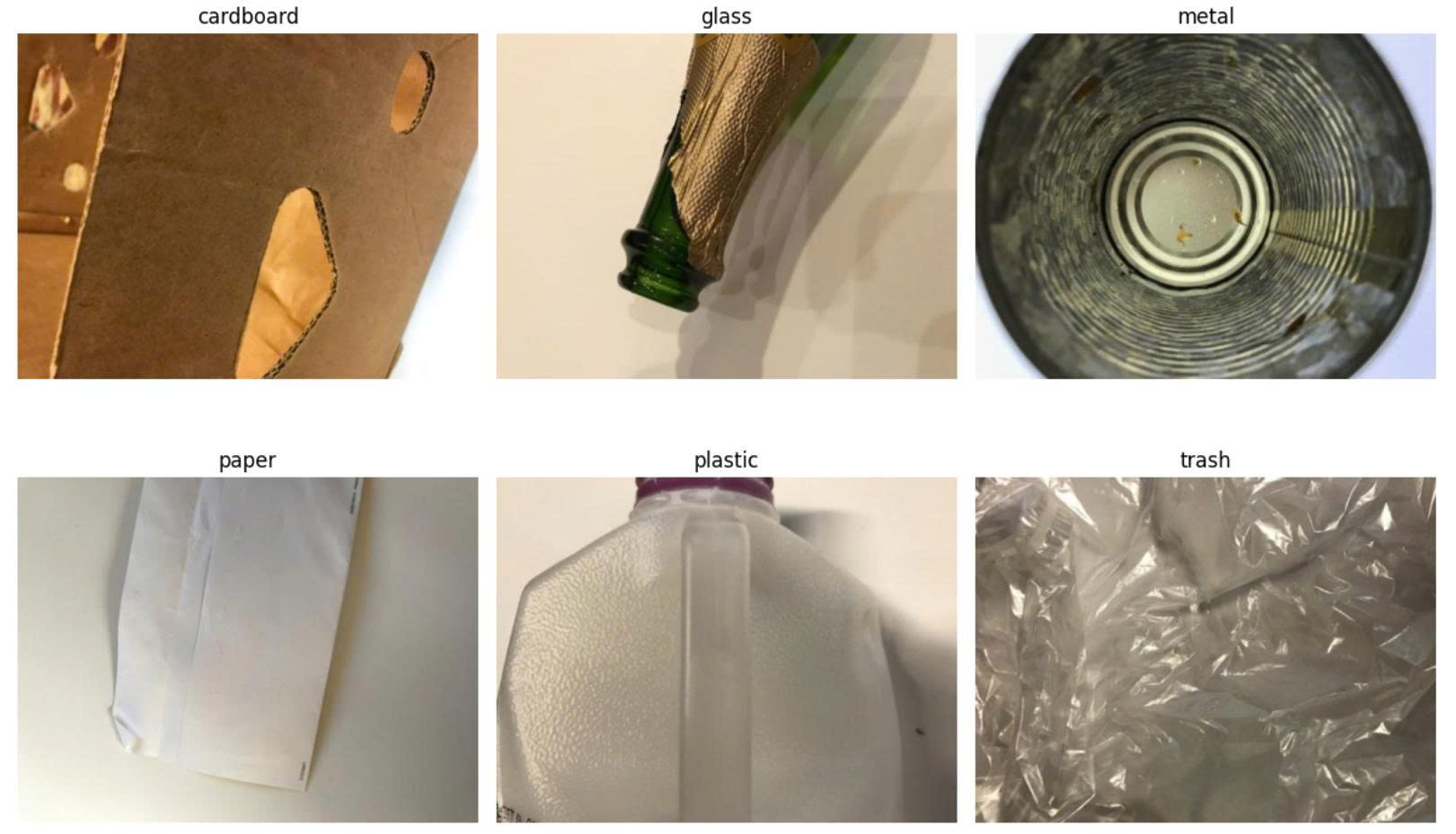}
    \caption{Sample images from the TrashNet dataset}
    \label{fig:trashnet}
\end{figure}

\subsection{Results and discussion}

This section presents a comprehensive evaluation of the proposed classification frameworks across the three selected datasets. To ensure a fair and consistent comparison, a uniform preprocessing pipeline was applied to all images prior to model training and evaluation. Specifically, each image was resized to a resolution of 400×400 pixels and subjected to data augmentation techniques designed to enhance model robustness and prevent overfitting.

Our analysis begins with a comparative evaluation of three distinct modeling paradigms, classical ML, DL, and our proposed hybrid approach, to establish the most effective framework. The detailed performance of each approach is presented in the following subsections.

\subsubsection{Performance of ML methods using handcrafted features}

The experimental results across the three datasets (Tables \ref{tab:classical_results_A}, \ref{tab:classical_results_B}, and \ref{tab:classical_results_trashnet}) 
 reveal consistent patterns regarding the effectiveness of classical ML with handcrafted features. Linear models such as LoR and SVM showed limited generalization capacity, with LoR consistently yielding the lowest performance (59--75\% accuracy), while SVM achieved a strong result only on dataset~B (82.40\%) but dropped notably on the others. Non-linear shallow models, including KNN and DT, also underperformed, with DT being the weakest overall (53--63\%), suggesting that handcrafted features alone do not provide sufficient discriminative power. In contrast, ensemble methods---particularly boosting algorithms---consistently emerged as the best-performing approaches. Both XGBoost and LightGBM achieved accuracies in the range of 78--85\% with balanced precision, recall, and F1-scores, clearly outperforming other models across all datasets. Notably, LightGBM attained the highest score on dataset~B (85.36\% accuracy, 85.29\% F1), while performance declined on dataset~C ($\approx$78\%), reflecting the increased difficulty of this dataset and the limited generalization ability of handcrafted features. Overall, these findings suggest that while handcrafted features combined with strong ensemble learners can yield moderate performance, they inevitably impose a ceiling. This motivates the transition to deep learning models, which are capable of automatically learning richer, more transferable feature representations from raw data.

\begin{table}[H]
\centering
\caption{Classification results of the classical ML approach on dataset A(\%)}
\label{tab:classical_results_A}
\begin{adjustbox}{max width=\textwidth}
\begin{tabular}{lcccc}
\toprule
\textbf{Model} & \textbf{Accuracy} & \textbf{Precision} & \textbf{Recall} & \textbf{F1-Score} \\
\midrule
SVM & 70.58 & 70.13 & 70.58 & 69.51 \\
RF & 74.89 & 75.13 & 74.89 & 73.85 \\
LoR & 66.42 & 66.24 & 66.42 & 66.26 \\
KNN & 62.34 & 62.19 & 62.34 & 61.92 \\
DT & 57.10 & 57.80 & 57.10 & 57.39 \\
XGBoost & \textbf{80.79} & \textbf{80.73} & \textbf{80.79} & \textbf{80.43} \\
LightGBM & 80.11 & 79.88 & 80.11 & 79.69 \\
\bottomrule
\end{tabular}
\end{adjustbox}
\end{table}

\begin{table}[H]
\centering
\caption{Classification results of the classical ML approach on dataset B(\%)}
\label{tab:classical_results_B}
\begin{adjustbox}{max width=\textwidth}
\begin{tabular}{lcccc}
\toprule
\textbf{Model} & \textbf{Accuracy} & \textbf{Precision} & \textbf{Recall} & \textbf{F1-Score} \\
\midrule
SVM & 82.40 & 82.12 & 82.40 & 82.19 \\
RF & 79.02 & 78.58 & 79.02 & 78.49 \\
LoR & 75.09 & 75.56 & 75.09 & 75.21 \\
KNN & 73.96 & 74.25 & 73.89 & 73.89 \\
DT & 63.16 & 63.58 & 63.16 & 63.33 \\
XGBoost & 84.79 & 84.69 & 84.79 & 84.62 \\
LightGBM & \textbf{85.36} & \textbf{85.36} & \textbf{85.37} & \textbf{85.29} \\
\bottomrule
\end{tabular}
\end{adjustbox}
\end{table}

\begin{table}[H]
\centering
\caption{Classification results of the classical ML approach on the dataset C(\%)}
\label{tab:classical_results_trashnet}
\begin{adjustbox}{max width=\textwidth}
\begin{tabular}{lcccc}
\toprule
\textbf{Model} & \textbf{Accuracy} & \textbf{Precision} & \textbf{Recall} & \textbf{F1-Score} \\
\midrule
SVM & 69.76 & 71.00 & 69.76 & 69.41 \\
RF & 70.55 & 71.82 & 70.55 & 69.93 \\
LoR & 59.88 & 60.72 & 59.88 & 60.16 \\
KNN & 63.64 & 63.85 & 63.64 & 63.58 \\
DT & 53.36 & 52.94 & 53.36 & 53.13 \\
XGBoost & \textbf{78.46} & \textbf{79.38} & \textbf{78.46} & \textbf{78.32} \\
LightGBM & 77.87 & 78.90 & 77.87 & 77.79 \\
\bottomrule
\end{tabular}
\end{adjustbox}
\end{table}

\subsubsection {Performance of deep learning models}

Tables~\ref{tab:dl_results_A}, \ref{tab:dl_results_B}, and \ref{tab:dl_results_trashnet} present the classification results of deep learning models across the three datasets. We analyze the performance per dataset, followed by general observations.

\textbf{Dataset A - Garbage Classification} \\
Images in dataset A exhibit high variability in background, lighting, and viewpoints, while the target objects occupy different proportions of the frame. Such diversity introduces background noise and increases class overlap. Consequently, DL models achieve high but not saturated performance: ResNet50 (94.23\%), EfficientNetV2S (96.61\%), ResNet101 (97.07\%), and customized ResNet50 (97.37\%). The customized version yields the best accuracy, slightly outperforming ResNet101 (+0.30\%) and EfficientNetV2S (+0.76\%). Compared to the standard ResNet50, it improves by +3.14\%, suggesting that the simplified head (GlobalAveragePooling2D + Dense) effectively suppresses irrelevant background features and aligns with the dataset’s class distribution. In contrast, the best ML model (XGBoost, 80.79\%) lags behind by 16.58\%, highlighting the limitations of handcrafted features under diverse acquisition conditions.

\begin{table}[H]
\centering
\caption{Classification results of the deep learning approach on dataset A(\%)}
\label{tab:dl_results_A}
\begin{adjustbox}{max width=\textwidth}
\begin{tabular}{lcccc}
\toprule
\textbf{Model} & \textbf{Accuracy} & \textbf{Precision} & \textbf{Recall} & \textbf{F1-Score} \\
\midrule
ResNet50 & 94.23 & 94.22 & 94.22 & 94.23 \\
ResNet101 & 97.07 & 97.09 & 97.05 & 97.06 \\
Customized ResNet50 & \textbf{97.37} & \textbf{97.44} & \textbf{97.27} & \textbf{97.33} \\
EfficientNetV2S & 96.61 & 96.62 & 96.61 & 96.61 \\
\bottomrule
\end{tabular}
\end{adjustbox}
\end{table}

\textbf{Dataset B - Household Garbage} \\
Images are more object-centric with relatively uniform backgrounds and stable acquisition conditions. This reduces intra-class variance, enabling transfer learning to leverage ImageNet features more effectively. Accordingly, DL models reach very high performance: ResNet50 (95.15\%), ResNet101 (97.59\%), and EfficientNetV2S (97.56\%). The customized ResNet50 nearly saturates performance with 99.74\% accuracy and 99.73\% F1, outperforming ResNet101 (+2.15\%) and EfficientNetV2S (+2.18\%). This suggests that the tailored head produces decision boundaries closely aligned with the dataset’s feature space. Compared to the best ML model (LightGBM, 85.36\%), the margin is +14.38\%.

\begin{table}[H]
\centering
\caption{Classification results of the deep learning approach on dataset B(\%)}
\label{tab:dl_results_B}
\begin{adjustbox}{max width=\textwidth}
\begin{tabular}{lcccc}
\toprule
\textbf{Model} & \textbf{Accuracy} & \textbf{Precision} & \textbf{Recall} & \textbf{F1-Score} \\
\midrule
ResNet50 & 95.15 & 95.13 & 95.15 & 95.10 \\
ResNet101 & 97.59 & 97.61 & 97.59 & 97.59 \\
Customized ResNet50 & \textbf{99.74} & \textbf{99.80} & \textbf{99.74} & \textbf{99.73}  \\
EfficientNetV2S & 97.56 & 97.58 & 97.56 & 97.56 \\
\bottomrule
\end{tabular}
\end{adjustbox}
\end{table}

\textbf{Dataset C - TrashNet} \\
This dataset is smaller in scale, with close-up images often containing specular highlights and variable viewpoints—factors that increase feature overlap between classes (e.g., glass vs.\ plastic). As a result, baseline DL performance is lower: ResNet50 (87.45\%) and EfficientNetV2S (91.37\%). However, scaling to deeper architectures (ResNet101, 97.59\%) and especially employing the customized head (97.63\%) restore robust performance. Notably, the customized ResNet50 achieves a +10.18\% improvement over the vanilla ResNet50, the largest margin among all datasets. This indicates that in low-data and noisy conditions, reducing the parameterization of the output head helps mitigate overfitting while preserving essential material cues. The best ML model (XGBoost, 78.46\%) remains far behind by +19.17\%.

\begin{table}[H]
\centering
\caption{Classification results of the deep learning approach on the dataset C(\%)}
\label{tab:dl_results_trashnet}
\begin{adjustbox}{max width=\textwidth}
\begin{tabular}{lcccc}
\toprule
\textbf{Model} & \textbf{Accuracy} & \textbf{Precision} & \textbf{Recall} & \textbf{F1-Score} \\
\midrule
ResNet50 & 87.45 & 87.80 & 87.45 & 87.48 \\
ResNet101 & 97.59 & 97.61 & 97.59 & 97.59 \\
Customized ResNet50 & \textbf{97.63} & \textbf{98.38} & \textbf{96.12} & \textbf{97.14} \\
EfficientNetV2S & 91.37 & 91.38 & 91.37 & 91.37 \\
\bottomrule
\end{tabular}
\end{adjustbox}
\end{table}

Overall, these results confirm two consistent patterns. First, the customized ResNet50 outperforms its vanilla counterpart across all datasets, with gains ranging from +3.14\% on dataset A to +10.18\% on dataset C. This shows that simplifying the classification head improves robustness to noise, class overlap, and limited training data. Second, deep learning models as a whole clearly surpass ML baselines based on handcrafted features, with margins of 14–19\% in accuracy. This performance gap reflects the stronger representational capacity of convolutional backbones to capture visual patterns beyond what hand-engineered descriptors can provide.

\subsubsection{Performance of ML methods using deep features (hybrid approach)}

Since customized ResNet50 proved to be the most effective deep learning model, it was subsequently selected as the feature extractor for our final hybrid approach. This framework proved to be exceptionally effective, yielding the highest performance across all three datasets and highlighting its effectiveness over the other paradigms.

Tables~\ref{tab:hybrid_results_A_custom}, \ref{tab:hybrid_results_B_custom}, and \ref{tab:hybrid_results_trashnet_custom} present the classification results obtained from the hybrid approach, where deep features extracted from the customized ResNet50 were used as input to various machine learning classifiers. Across all three datasets, this strategy consistently yielded superior performance compared to standalone deep learning models.

For dataset A (Table~\ref{tab:hybrid_results_A_custom}), LoR and SVM achieved the highest results, with accuracy, precision, recall, and F1-score all reaching 99.87\%. LightGBM also performed competitively at 99.65\%. Other classifiers, including RF, KNN, DT, and XGBoost, obtained slightly lower scores (ranging from 97\% to 99.5\%). Notably, when compared with the best deep learning model on this dataset (97.37\% with customized ResNet50), the hybrid approach provided a clear improvement of approximately +2.5\%, highlighting the effectiveness of combining DL-based feature extraction with ML classifiers.

\begin{table}[H]
\centering
\caption{Classification results of the hybrid approach using customized ResNet50 features on dataset A(\%)}
\label{tab:hybrid_results_A_custom}
\begin{adjustbox}{max width=\textwidth}
\begin{tabular}{lcccc}
\toprule
\textbf{Model} & \textbf{Accuracy} & \textbf{Precision} & \textbf{Recall} & \textbf{F1-Score} \\
\midrule
LoR & \textbf{99.87} & \textbf{99.87} & \textbf{99.87} & \textbf{99.87} \\
RF & 99.47 & 99.48 & 99.47 & 99.47 \\
SVM & \textbf{99.87} & \textbf{99.87} & \textbf{99.87} & \textbf{99.87} \\
KNN & 97.98 & 98.01 & 97.98 & 97.98 \\
DT & 97.37 & 97.38 & 97.37 & 97.37 \\
XGBoost & 99.60 & 99.60 & 99.60 & 99.60 \\
LightGBM & 99.65 & 99.66 & 99.65 & 99.66 \\
\bottomrule
\end{tabular}
\end{adjustbox}
\end{table}

On dataset B (Table~\ref{tab:hybrid_results_B_custom}), the hybrid approach delivered near-perfect performance. LoR once again outperformed all other methods, achieving 99.97--99.99\% across all metrics. RF, SVM, and LightGBM followed closely with results around or above 99.8\%. While DT and KNN were relatively lower (98--99\%), they still outperformed their counterparts trained on handcrafted features. Compared to the deep learning baseline (99.74\% with customized ResNet50), the hybrid method further refined the classification results, achieving virtually error-free outcomes.

\begin{table}[H]
\centering
\caption{Classification results of the hybrid approach using customized ResNet50 features on dataset B(\%)}
\label{tab:hybrid_results_B_custom}
\begin{adjustbox}{max width=\textwidth}
\begin{tabular}{lcccc}
\toprule
\textbf{Model} & \textbf{Accuracy} & \textbf{Precision} & \textbf{Recall} & \textbf{F1-Score} \\
\midrule
RF & 99.77 & 99.71 & 99.81 & 99.76 \\
LoR & \textbf{99.97} & \textbf{99.98} & \textbf{99.99} & \textbf{99.99} \\
SVM  & 99.94 & 99.90 & 99.94 & 99.92 \\
KNN & 99.74 & 99.67 & 99.76 & 99.72 \\
DT & 98.84 & 98.37 & 98.38 & 98.37 \\
XGBoost & 99.81 & 99.77 & 99.81 & 99.79 \\
LightGBM & 99.87 & 99.78 & 99.83 & 99.81 \\
\bottomrule
\end{tabular}
\end{adjustbox}
\end{table}

The most striking results are observed in dataset C (Table~\ref{tab:hybrid_results_trashnet_custom}), where LoR, RF, SVM, and LightGBM all attained 100\% accuracy, precision, recall, and F1-score, thereby eliminating classification errors entirely. Although XGBoost also performed strongly (99.8--99.9\%), KNN and DT lagged behind (97--99\%). Considering that the best deep learning model on dataset C reached only 97.63\%, the hybrid approach demonstrates a remarkable improvement of +2--3\%, underscoring its robustness and generalizability across challenging datasets.

\begin{table}[H]
\centering
\caption{Classification results of the hybrid approach using customized ResNet50 features on the dataset C(\%)}
\label{tab:hybrid_results_trashnet_custom}
\begin{adjustbox}{max width=\textwidth}
\begin{tabular}{lcccc}
\toprule
\textbf{Model} & \textbf{Accuracy} & \textbf{Precision} & \textbf{Recall} & \textbf{F1-Score} \\
\midrule
RF & \textbf{100.00} & \textbf{100.00} & \textbf{100.00} & \textbf{100.00} \\
LoR & \textbf{100.00} & \textbf{100.00} & \textbf{100.00} & \textbf{100.00} \\
SVM & \textbf{100.00} & \textbf{100.00} & \textbf{100.00} & \textbf{100.00} \\
KNN & 98.82 & 98.93 & 99.00 & 98.96 \\
DT & 97.63 & 97.44 & 97.00 & 97.20 \\
XGBoost & 99.80 & 99.49 & 99.87 & 99.68 \\
LightGBM & \textbf{100.00} & \textbf{100.00} & \textbf{100.00} & \textbf{100.00} \\
\bottomrule
\end{tabular}
\end{adjustbox}
\end{table}

\begin{figure}[H]
    \centering
    \includegraphics[width=\textwidth]{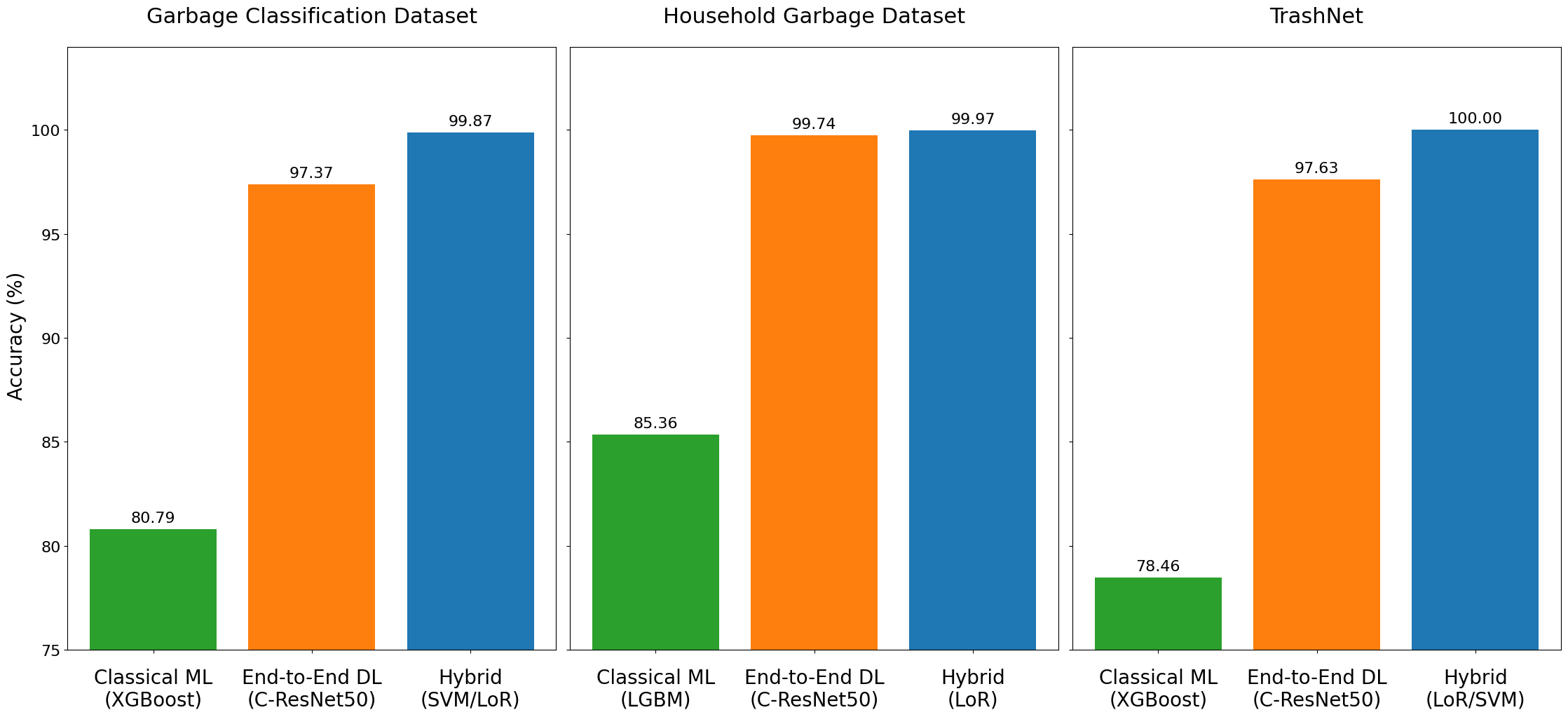}
    \caption{Summary of the best performance from each approach across all datasets.}
    \label{fig:initial_comparison_summary_final}
\end{figure}

The superior performance of the hybrid DL-ML approach, visually summarized in Figure~\ref{fig:initial_comparison_summary_final}, can be attributed to several key factors. First, our customized ResNet50 serves as a powerful feature extractor, generating highly discriminative representations that vastly outperform handcrafted features. As shown across all three datasets, the performance gap between the classical ML approach and the two deep learning-based paradigms is substantial.

Second, the results reveal the consistent advantage of the hybrid framework over the end-to-end approach. By separating feature learning (DL) from classification (ML), the hybrid method allows for the selection of a classical classifier best suited for the feature space. Our findings suggest that the deep features extracted are nearly linearly separable, which explains why simpler classifiers like LoR and SVM consistently achieve the best results, often surpassing the performance of the DL model's dense layers.

Finally, the synergy between a fine-tuned backbone and a well-matched ML classifier leads to state-of-the-art performance. The hybrid strategy consistently leverages the complementary strengths of both paradigms, leading to superior and more stable results across different data distributions, as evidenced by its top performance on all three benchmarks.

\subsection{Dataset integrity analysis and correction}
\begin{figure}[H]
    \centering
    \includegraphics[width=1\textwidth]{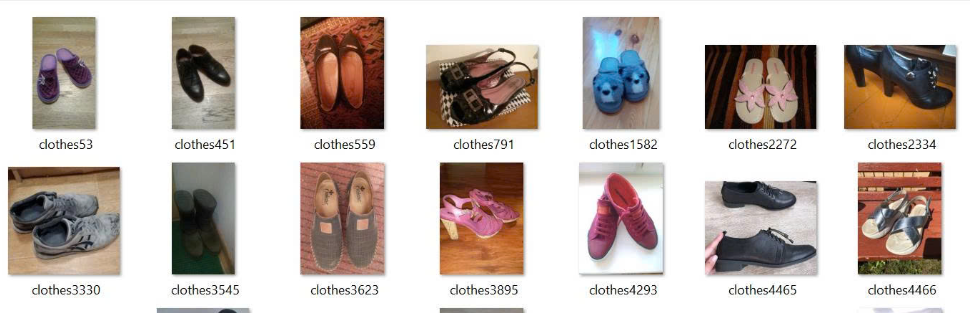}
    \caption{An example of a mislabeled image found during the inspection.}
    \label{fig:mislabeled_example}
\end{figure}

During the initial analysis, we discovered a considerable number of mislabeled images in the Household Garbage dataset. To enhance the reliability of subsequent evaluations, a manual inspection and correction were conducted, resulting in the identification and relabeling of 43 images. An illustrative example of such mislabeling is shown in Figure~\ref{fig:mislabeled_example}.

To assess the impact of data quality issues, we conducted a focused evaluation of the ResNet50 architecture, comparing both deep learning models and the hybrid approach on the original and corrected versions of the Household Garbage dataset. The results, summarized in Table~\ref{tab:household_resnet50_impact}, highlight distinct differences in how these paradigms respond to label noise.

On the original dataset, our hybrid framework, ResNet50 combined with LoR, achieved 99.97\% accuracy, surpassing the 99.87\% reported by \citep{celik} and establishing a new state-of-the-art. This result underscores the robustness of the hybrid strategy, even when trained on mislabeled data.

Following label correction, the performance gap between the two approaches became more pronounced. The hybrid approach further improved, with the ResNet50 and SVM variant reaching 100\% accuracy. In contrast, the deep learning model baseline dropped from 99.74\% on the original dataset to 99.16\% after correction, suggesting that it had overfitted to the label noise.

Taken together, these findings confirm the superior robustness of the hybrid approach, with its 100\% accuracy on the corrected dataset serving as the most reliable benchmark for this task.

\begin{table}[H]
\centering
\caption{Performance comparison of ResNet50-based models on the original and corrected Household Garbage dataset(\%)}
\label{tab:household_resnet50_impact}
\begin{adjustbox}{max width=\textwidth}
\begin{tabular}{llcccc}
\toprule
\textbf{Model } & \textbf{Dataset version} & \textbf{Accuracy} & \textbf{Precision} & \textbf{Recall} & \textbf{F1-Score} \\
\midrule
Customized ResNet50  & Original & 99.74 & 99.74 & 99.74 & 99.74 \\
Customized ResNet50  & Corrected & 99.16 & 98.85 & 99.08 & 98.96 \\
\midrule
Customized ResNet50 + LoR & Original & 99.97 & 99.98 & 99.99 & 99.99 \\
Customized ResNet50 + SVM & Corrected & \textbf{100} & \textbf{100} & \textbf{100} & \textbf{100} \\
\bottomrule
\end{tabular}
\end{adjustbox}
\end{table}

\subsection{Impact of feature selection on deep features}

In the hybrid approach, we utilize a 2048-dimensional feature vector extracted via Global Average Pooling from the customized ResNet50 backbone. Although this significantly improved classification performance, the feature set was still excessively large, potentially causing redundancy and high computational cost. Therefore, we applied feature selection to reduce the number of input features while maintaining, or even enhancing, the model’s accuracy.

\textbf{Dataset A - Garbage Classification} 

As an initial step, we selected the top 100 features out of the 2048 extracted ones, since this represents a substantial dimensionality reduction (95\%) while still being large enough to retain the most informative signals and provide a reasonable baseline for further comparison. After observing strong performance with 100 features, we gradually reduced the number of selected features (90, 80, 70, and 50) to identify the optimal threshold, rather than applying an overly aggressive reduction from the beginning. 

 The results in Table~\ref{tab:fs_impact_gc} clearly demonstrate the strong impact of feature selection on model performance. When reducing the number of features from 100 to 50, most classifiers maintained very high accuracy, and several even achieved near-perfect or perfect scores with only a small subset of features. The highest accuracy of 99.44\% was achieved by the LightGBM model using only the top 60 features, closely followed by XGBoost and RF at the 70--90 feature range with 99.39\%. Interestingly, performance slightly decreased at the 100-feature mark, suggesting that the most discriminative information is concentrated in a compact subset of features, while adding more may introduce redundancy or noise. SVM and LoR also performed remarkably well, achieving perfect accuracy but requiring slightly larger subsets (80--100 features), whereas KNN and DT remained the weakest, ranging between 96\% and 98\%. Overall, these findings confirm that the proposed feature selection pipeline enables a massive dimensionality reduction of about 96--97\% (from 2048 to as few as 60--100 features) while preserving or even enhancing classification accuracy, thereby ensuring robustness, efficiency, and deployability of the hybrid pipeline.

\begin{table}[htbp]
\centering
\caption{Impact of feature quantity on model performance on Garbage Classification dataset(\%)}
\label{tab:fs_impact_gc}
\resizebox{!}{0.50\textheight}{
\begin{tabular}{l l cccc}
\toprule
\textbf{Features used} & \textbf{Classifier} & \textbf{Accuracy} & \textbf{Precision} & \textbf{Recall} & \textbf{F1-Score} \\
\midrule
\multirow{7}{*}{Top 100}
 & LoR & 99.18 & 99.18 & 99.18 & 99.18 \\
 & RF & 99.28 & 99.29 & 99.28 & 99.28 \\
 & SVM  & 99.20 & 99.20 & 99.20 & 99.20 \\
 & KNN & 97.62 & 97.66 & 97.62 & 97.63 \\
 & DT & 97.07 & 97.09 & 97.07 & 97.07 \\
 & XGBoost & 98.91 & 98.92 & 98.91 & 98.91 \\
 & LightGBM & 98.56 & 98.57 & 98.56 & 98.56 \\
\midrule
\multirow{7}{*}{Top 90}
 & LoR & 98.99 & 99.00 & 98.99 & 98.99 \\
 & RF & 99.39 & 99.39 & 99.39 & 99.39 \\
 & SVM  & 99.18 & 99.18 & 99.18 & 99.18 \\
 & KNN & 97.40 & 97.43 & 97.40 & 97.40 \\
 & DT & 97.98 & 97.98 & 97.98 & 97.98 \\
 & XGBoost & 99.39 & 99.39 & 99.39 & 99.39 \\
 & LightGBM & 99.39 & 99.39 & 99.39 & 99.39 \\
\midrule
\multirow{7}{*}{Top 80}
 & LoR & 98.80 & 98.82 & 98.80 & 98.81 \\
 & RF & 99.18 & 99.18 & 99.18 & 99.18 \\
 & SVM  & 98.96 & 98.96 & 98.96 & 98.96 \\
 & KNN & 97.08 & 97.11 & 97.08 & 97.08 \\
 & DT & 97.66 & 97.67 & 97.66 & 97.66 \\
 & XGBoost & 99.23 & 99.24 & 99.23 & 99.23 \\
 & LightGBM & 99.28 & 99.29 & 99.28 & 99.28 \\
\midrule
\multirow{7}{*}{Top 70}
 & LoR & 98.51 & 98.52 & 98.51 & 98.51 \\
 & RF & 99.26 & 99.26 & 99.26 & 99.26 \\
 & SVM  & 98.91 & 98.91 & 98.91 & 98.91 \\
 & KNN & 97.10 & 97.12 & 97.10 & 97.10 \\
 & DT & 97.74 & 97.75 & 97.74 & 97.74 \\
 & XGBoost & 99.39 & 99.39 & 99.39 & 99.39 \\
 & LightGBM & 99.36 & 99.36 & 99.36 & 99.36 \\
\midrule
\multirow{7}{*}{Top 60}
 & LoR & 98.14 & 98.15 & 98.14 & 98.14 \\
 & RF & 99.15 & 99.15 & 99.15 & 99.15 \\
 & SVM  & 98.72 & 98.73 & 98.72 & 98.72 \\
 & KNN & 96.63 & 96.65 & 96.63 & 96.62 \\
 & DT & 97.13 & 97.14 & 97.13 & 97.13 \\
 & XGBoost & 99.31 & 99.31 & 99.31 & 99.31 \\
 & LightGBM & \textbf{99.44} & \textbf{99.44} & \textbf{99.44} & \textbf{99.44} \\
\midrule
\multirow{7}{*}{Top 50}
 & LoR & 97.10 & 97.13 & 97.10 & 97.11 \\
 & RF & 98.86 & 98.86 & 98.86 & 98.86 \\
 & SVM  & 97.50 & 97.51 & 97.50 & 97.50 \\
 & KNN & 95.91 & 95.98 & 95.91 & 95.92 \\
 & DT & 97.10 & 97.10 & 97.10 & 97.10 \\
 & XGBoost & 99.23 & 99.23 & 99.23 & 99.23 \\
 & LightGBM & 99.18 & 99.18 & 99.18 & 99.18 \\
\bottomrule
\end{tabular}
}
\end{table}

\vspace{0.2cm}
\textbf{Dataset B - Household Garbage } 

The results in Table~\ref{tab:fs_impact_hgd} replicate and extend the findings obtained from the  Garbage Classification dataset, confirming the robustness and consistency of the proposed feature selection pipeline. Peak performance is concentrated within the top 100 features: SVM achieved the highest accuracy of 99.87\% using only the top 90 features, while RF likewise reached up to 99.84\% with 90--100 features. Boosting methods (XGBoost and LightGBM) exhibited remarkable stability, delivering consistently high accuracies (99.7--99.8\%) even when the feature set was reduced to 50--70 features, which indicates robustness to aggressive dimensionality reduction. LoR produced competitive scores (99.7--99.8\%) at the larger subsets (80--100), whereas KNN and DT were more sensitive to reduction, dropping to the high 97\%--98\% range for smaller subsets. Overall, these results show that a massive reduction in dimensionality (over 95\%, from 2048 down to 90 features) is feasible on this complex dataset while still achieving near–state-of-the-art performance; this strongly supports the generalizability and efficiency of our proposed feature selection pipeline. The relatively small fluctuations in performance across adjacent subset sizes (50--100) suggest a broad plateau of near-optimal feature sets rather than a single sharp optimum, implying that practitioners may choose slightly larger subsets (e.g., 80--100) to increase robustness with minimal cost to parsimony.

\textbf{Dataset C - TrashNet} 
    
Finally, to confirm the generalizability of our approach, the experiment was repeated on the TrashNet dataset with a fine-grained analysis of feature subsets ranging from 100 down to 50.
The results in Table~\ref{tab:fs_impact_trashnet} clearly demonstrate the strong impact of feature selection on model performance. When reducing the number of features from 100 down to 50, most classifiers maintained very high accuracy, and several even achieved perfect scores with only a small subset of features. This indicates that more than 90\% of the original 2048 deep features are redundant or contribute little to the final classification task. Among the classifiers, XGBoost and LightGBM stand out by reaching 100\% accuracy with only 70 selected features and sustaining this performance with larger subsets, confirming their strong capability to exploit the most relevant information even from highly compact representations. SVM and LoR also achieved perfect accuracy but required slightly larger subsets (80--100 features), which is remarkable given their sensitivity to high dimensionality and redundant data. RF consistently performed well (around 99.6--99.8\%) but did not reach 100\%, suggesting it is less optimal when the feature space is aggressively reduced. In contrast, KNN and DT showed the weakest results, ranging from 96\% to 98\%, reflecting their reliance on distance measures or splitting rules that are more affected by reduced representations. Overall, these findings confirm that the proposed feature selection pipeline is highly effective, as it not only reduces the dimensionality by approximately 96--97\% (from 2048 to as few as 70--100 features) but also preserves or even enhances classification accuracy, leading to lighter, more efficient, and more deployable models.

\begin{table}[htbp]
\centering
\caption{Impact of feature quantity on model performance on corrected Household Garbage dataset(\%)}
\label{tab:fs_impact_hgd}
\resizebox{!}{0.50\textheight}{
\begin{tabular}{l l cccc}
\toprule
\textbf{Features used} & \textbf{Classifier} & \textbf{Accuracy} & \textbf{Precision} & \textbf{Recall} & \textbf{F1-Score} \\
\midrule
\multirow{7}{*}{Top 100}
 & RF & 99.84 & 99.78 & 99.78 & 99.78 \\
 & LoR & 99.84 & 99.83 & 99.80 & 99.82 \\
 & SVM  & 99.81 & 99.82 & 99.79 & 99.80 \\
 & DT & 98.93 & 98.46 & 98.43 & 98.44 \\
 & KNN & 99.16 & 98.73 & 98.74 & 98.73 \\
 & XGBoost & 99.77 & 99.69 & 99.68 & 99.68 \\
 & LightGBM & 99.77 & 99.71 & 99.67 & 99.69 \\
\midrule
\multirow{7}{*}{Top 90}
 & RF & 99.84 & 99.83 & 99.76 & 99.80 \\
 & LoR & 99.74 & 99.66 & 99.61 & 99.63 \\
 & SVM  & \textbf{99.87} & \textbf{99.93} & \textbf{99.90} & \textbf{99.91} \\
 & DT & 98.87 & 98.26 & 98.33 & 98.30 \\
 & KNN & 99.13 & 98.65 & 98.69 & 98.67 \\
 & XGBoost & 99.74 & 99.68 & 99.62 & 99.65 \\
 & LightGBM & 99.71 & 99.61 & 99.57 & 99.59 \\
\midrule
\multirow{7}{*}{Top 80}
 & RF & 99.77 & 99.71 & 99.67 & 99.69 \\
 & LoR & 99.68 & 99.57 & 99.56 & 99.56 \\
 & SVM  & 99.81 & 99.82 & 99.79 & 99.81 \\
 & DT & 98.80 & 98.19 & 98.31 & 98.25 \\
 & KNN & 99.00 & 98.47 & 98.47 & 98.46 \\
 & XGBoost & 99.77 & 99.72 & 99.68 & 99.70 \\
 & LightGBM & 99.74 & 99.68 & 99.63 & 99.66 \\
\midrule
\multirow{7}{*}{Top 70}
 & RF & 99.74 & 99.63 & 99.62 & 99.63 \\
 & LoR & 99.48 & 99.22 & 99.23 & 99.22 \\
 & SVM  & 99.64 & 99.51 & 99.52 & 99.52 \\
 & DT & 98.67 & 98.09 & 98.04 & 98.06 \\
 & KNN & 98.77 & 98.09 & 98.11 & 98.09 \\
 & XGBoost & 99.77 & 99.73 & 99.68 & 99.70 \\
 & LightGBM & 99.74 & 99.60 & 99.65 & 99.62 \\
\midrule
\multirow{7}{*}{Top 60}
 & RF & 99.77 & 99.69 & 99.67 & 99.68 \\
 & LoR & 99.55 & 99.38 & 99.37 & 99.37 \\
 & SVM  & 99.42 & 99.18 & 99.15 & 99.16 \\
 & DT & 98.48 & 97.68 & 97.72 & 97.70 \\
 & KNN & 98.45 & 97.56 & 97.65 & 97.58 \\
 & XGBoost & 99.74 & 99.66 & 99.62 & 99.64 \\
 & LightGBM & 99.74 & 99.56 & 99.63 & 99.59 \\
\midrule
\multirow{7}{*}{Top 50}
 & RF & 99.55 & 99.28 & 99.30 & 99.28 \\
 & LoR & 98.87 & 98.31 & 98.32 & 98.31 \\
 & SVM  & 99.13 & 98.73 & 98.74 & 98.72 \\
 & DT & 97.96 & 96.81 & 97.08 & 96.94 \\
 & KNN & 98.38 & 97.39 & 97.52 & 97.40 \\
 & XGBoost & 99.64 & 99.54 & 99.48 & 99.51 \\
 & LightGBM & 99.74 & 99.59 & 99.62 & 99.61 \\
\bottomrule
\end{tabular}
}
\end{table}

\begin{table}[htbp]
\centering
\caption{Impact of feature quantity on model performance on the TrashNet dataset(\%)}
\label{tab:fs_impact_trashnet}
\resizebox{!}{0.50\textheight}{
\begin{tabular}{l l cccc}
\toprule
\textbf{Features used} & \textbf{Classifier} & \textbf{Accuracy} & \textbf{Precision} & \textbf{Recall} & \textbf{F1-Score} \\
\midrule
\multirow{7}{*}{Top 100} 
 & RF & 99.80 & 99.84 & 99.85 & 99.84 \\
 & LoR & \textbf{100.00} & \textbf{100.00} & \textbf{100.00} & \textbf{100.00} \\
 & SVM & \textbf{100.00} & \textbf{100.00} & \textbf{100.00} & \textbf{100.00} \\
 & DT & 98.03 & 97.65 & 97.98 & 97.81 \\
 & KNN & 98.03 & 97.87 & 98.53 & 98.15 \\
 & XGBoost & \textbf{100.00} & \textbf{100.00} & \textbf{100.00} & \textbf{100.00} \\
 & LightGBM & \textbf{100.00} & \textbf{100.00} & \textbf{100.00} & \textbf{100.00} \\
\midrule
\multirow{7}{*}{Top 90} 
 & RF & 99.61 & 99.33 & 99.72 & 99.52 \\
 & LoR & \textbf{100.00} & \textbf{100.00} & \textbf{100.00} & \textbf{100.00} \\
 & SVM & \textbf{100.00} & \textbf{100.00} & \textbf{100.00} & \textbf{100.00} \\
 & DT & 98.42 & 97.89 & 98.65 & 98.25 \\
 & KNN & 97.63 & 97.54 & 98.07 & 97.77 \\
 & XGBoost & 99.80 & 99.49 & 99.71 & 99.60 \\
 & LightGBM & \textbf{100.00} & \textbf{100.00} & \textbf{100.00} & \textbf{100.00} \\
\midrule
\multirow{7}{*}{Top 80} 
 & RF & 99.61 & 99.33 & 99.72 & 99.52 \\
 & LoR & 99.80 & 99.49 & 99.87 & 99.68 \\
 & SVM & \textbf{100.00} & \textbf{100.00} & \textbf{100.00} & \textbf{100.00} \\
 & DT & 97.83 & 97.13 & 97.54 & 97.31 \\
 & KNN & 97.63 & 97.04 & 98.07 & 97.49 \\
 & XGBoost & 99.80 & 99.49 & 99.71 & 99.60 \\
 & LightGBM & \textbf{100.00} & \textbf{100.00} & \textbf{100.00} & \textbf{100.00} \\
\midrule
\multirow{7}{*}{Top 70} 
 & RF & 99.61 & 99.33 & 99.72 & 99.52 \\
 & LoR & 99.80 & 99.49 & 99.87 & 99.68 \\
 & SVM  & 99.41 & 99.23 & 99.57 & 99.39 \\
 & DT & 97.63 & 96.76 & 97.10 & 96.89 \\
 & KNN & 97.24 & 96.33 & 97.76 & 96.96 \\
 & XGBoost & \textbf{100.00} & \textbf{100.00} & \textbf{100.00} & \textbf{100.00} \\
 & LightGBM & \textbf{100.00} & \textbf{100.00} & \textbf{100.00} & \textbf{100.00} \\
\midrule
\multirow{7}{*}{Top 60} 
 & RF & 99.80 & 99.49 & 99.87 & 99.68 \\
 & LoR & 99.41 & 98.94 & 99.57 & 99.24 \\
 & SVM  & 99.41 & 98.94 & 99.57 & 99.24 \\
 & DT & 97.04 & 96.38 & 96.63 & 96.47 \\
 & KNN & 97.04 & 96.35 & 97.61 & 96.90 \\
 & XGBoost & 99.80 & 99.49 & 99.84 & 99.66 \\
 & LightGBM & 99.80 & 99.72 & 99.85 & 99.78 \\
\midrule
\multirow{7}{*}{Top 50} 
 & RF & 99.80 & 99.49 & 99.87 & 99.68 \\
 & LoR & 99.41 & 99.06 & 99.55 & 99.30 \\
 & SVM  & 99.01 & 98.57 & 99.27 & 98.91 \\
 & DT & 98.03 & 96.82 & 97.25 & 97.02 \\
 & KNN & 97.04 & 96.00 & 97.75 & 96.78 \\
 & XGBoost & 99.80 & 99.49 & 99.84 & 99.66 \\
 & LightGBM & 99.80 & 99.49 & 99.84 & 99.66 \\
\bottomrule
\end{tabular}
}
\end{table}

From the results presented in the three tables, it is evident that the proposed hybrid framework offers two major advantages: a drastic reduction in data dimensionality and a significant gain in computational efficiency. Across all datasets, we consistently observed that a compact subset of the most informative features (typically within the Top 50--100 range, representing over 95\% reduction from the original 2048 features) was sufficient to achieve, and in some cases even surpass, the accuracy of the full feature set. This reduction not only eliminates redundancy but also accelerates both training and inference.

\subsection{Computation time analysis}

In addition to comparing classification performance across approaches, ranging from handcrafted features to deep learning models and combined handcrafted and deep features, we also compare the pipelines in terms of computation time. The focus of this section is on training duration, feature extraction cost, and per-sample inference latency, as these metrics illuminate the practical trade-offs involved in deployment. To keep the analysis concise and representative, we do not report every possible method; instead, we (i) use the customized ResNet50 model as a representative DL model, and (ii) report hybrid results only for the top-performing ML classifiers identified in the previous hybrid evaluation. This selective presentation is intentional and provides a meaningful comparison without redundant detail.

It is important to note that in the hybrid pipeline, feature extraction is performed after the CNN has been trained. This means that the training cost of the CNN is a prerequisite for both end-to-end and hybrid pipelines. The difference lies in how inference is performed: the DL model relies entirely on the CNN for each prediction, whereas in the hybrid approach, a one-time feature extraction is carried out across the training dataset, and during inference each new sample still undergoes CNN-based feature extraction but the subsequent classification is delegated to lightweight ML models.

\textit{Note on measurement:} Feature-extraction latency is measured as a single forward pass of the ResNet50 backbone (with GAP) in inference mode, averaged over the test set with identical batch size and data pipeline across datasets. This ensures that latency values are directly comparable across datasets and between end-to-end and hybrid pipelines.


\textbf{Dataset A - Garbage Classification} 

As a starting point, we analyze the computation time on the largest dataset. Table \ref{tab:speed_summary_garbage} shows that the customized ResNet50 required 7064.41 seconds for training and achieved an inference latency of 12.67 ms per sample. In the hybrid pipeline, after CNN training a one-time feature extraction over the dataset incurred 140.62 seconds, while training the classifiers took 8.09 seconds for SVM and 122.12 seconds for LoR, giving total training times of 7213.12 and 7327.15 seconds, respectively. At inference, each new sample still required CNN-based feature extraction ($\approx 9.34$ ms) followed by classification, leading to total latencies of 10.11 ms (SVM) and 9.36 ms (LoR), faster than the end-to-end baseline (12.67 ms) by a few milliseconds.

\begin{table}[h]
\centering
\begin{threeparttable}
\caption{Computation time analysis on the Garbage Classification dataset}
\label{tab:speed_summary_garbage}
\begin{tabular}{lccc}
\hline
\textbf{Model } & \textbf{Training (s)} & \textbf{FE (s)} & \textbf{Inference (ms/sample)} \\ 
\hline
Customized ResNet50        & 7064.41 & -      & 12.67 \\
Hybrid (using SVM)         & 7213.12 & 140.62 & 10.11 \\
Hybrid (using LoR)         & 7327.15 & 140.62 & 9.36 \\
\hline
\end{tabular}
\begin{tablenotes}
\item \textit{Note:} Training time of hybrid models = Training customized ResNet50 + Feature extraction + Training ML methods. 

Inference time of hybrid models = Feature extraction + Inference (SVM/LoR).
\end{tablenotes}
\end{threeparttable}
\end{table}

\textbf{Dataset B - Household Garbage}

Table \ref{tab:speed_summary_household} shows that the customized ResNet50 required 5243.63 seconds for training and achieved an inference latency of 6.32 ms per sample. In the hybrid pipeline, after CNN training a one-time feature extraction over the dataset incurred 76.55 seconds, while training the classifiers took 4.94 seconds for SVM and 89.35 seconds for LoR, giving total training times of 5325.12 and 5409.53 seconds, respectively. 

At inference, each new sample still required CNN-based feature extraction ($\approx 6.31$ ms) followed by classification, resulting in total latencies of 6.98 ms (SVM) and 6.33 ms (LoR). Overall differences are only fractions of a millisecond compared to the end-to-end baseline, but the negligible classifier overhead remains attractive, especially when features are reused across multiple lightweight classifiers.

\begin{table}[H]
\centering
\begin{threeparttable}
\caption{Computation time analysis on the Household Garbage dataset}
\label{tab:speed_summary_household}
\begin{tabular}{lccc}
\hline
\textbf{Model } & \textbf{Training (s)} & \textbf{FE (s)} & \textbf{Inference (ms/sample)} \\ 
\hline
Customized ResNet50        & 5243.63 & -     & 6.32 \\
Hybrid (using SVM)         & 5325.12 & 76.55 & 6.98 \\
Hybrid (using LoR)         & 5409.53 & 76.55 & 6.33 \\
\hline
\end{tabular}
\end{threeparttable}
\end{table}

\textbf{Dataset C - TrashNet}

Table \ref{tab:speed_summary_trashnet} shows that the customized ResNet50 required 1373.07 seconds for training and achieved an inference latency of 9.97 ms per sample. In the hybrid pipeline, after CNN training a one-time feature extraction was performed on the training dataset (18.14 seconds), and at inference each sample still required CNN-based feature extraction ($\approx 9.07$ ms) before classification by lightweight models. This design reduces classifier overhead to 0.21 ms (SVM) and 0.02 ms (LoR), yielding total latencies of 9.28 ms and 9.09 ms per sample, respectively—a modest reduction (sub-millisecond) compared to the end-to-end baseline. The benefit, therefore, lies primarily in the negligible classifier cost rather than a large drop in overall per-sample latency.
Table \ref{tab:speed_summary_trashnet} shows that the customized ResNet50 required 1373.07 seconds for training and achieved an inference latency of 9.97 ms per sample. In the hybrid pipeline, after CNN training a one-time feature extraction was performed on the training dataset (18.14 seconds), and at inference each sample still required CNN-based feature extraction ($\approx 9.07$ ms) before classification by lightweight models. This design reduces classifier overhead to 0.21 ms (SVM) and 0.02 ms (LoR), yielding total latencies of 9.28 ms and 9.09 ms per sample, respectively—a modest reduction (sub-millisecond) compared to the end-to-end baseline. The benefit, therefore, lies primarily in the negligible classifier cost rather than a large drop in overall per-sample latency.

\begin{table}[h]
\centering
\begin{threeparttable}
\caption{Computation time analysis on the TrashNet dataset}
\label{tab:speed_summary_trashnet}
\begin{tabular}{lccc}
\hline
\textbf{Model } & \textbf{Training (s)} & \textbf{FE (s)} & \textbf{Inference (ms/sample)} \\ 
\hline
Customized ResNet50        & 1373.07 & -     & 9.97 \\
Hybrid (using SVM)         & 1373.45 & 18.14 & 9.28 \\
Hybrid (using LoR)         & 1380.82 & 18.14 & 9.09 \\
\hline
\end{tabular}
\end{threeparttable}
\end{table}

Across the three datasets, several consistent patterns can be observed:
\begin{itemize}
\item \textbf{Training cost:} customized ResNet50 models require from roughly 1.4k to 7.1k seconds of training, whereas the additional training of hybrid classifiers (SVM or LoR) takes seconds to a few minutes.
\item \textbf{Feature extraction:} the hybrid pipeline introduces a CNN-based feature extraction stage that dominates inference cost, averaging approximately 6–10 ms per sample across datasets.
\item \textbf{Inference latency:} after feature extraction, hybrid ML classifiers operate at sub-millisecond latency, consistently faster than the DL model in the classification step; the net per-sample latency difference is typically only a few milliseconds.
\end{itemize}
In general, inference in the end-to-end pipeline requires the CNN to process each input fully and directly output predictions, tying latency to the deep model. By contrast, the hybrid pipeline separates feature extraction (CNN) from classification (lightweight ML), shifting most of the cost to the CNN pass. Although absolute differences are small (a few milliseconds), they are reproducible and can matter in high-throughput or edge deployments, particularly when features are reused or multiple classifiers share the same features.

Taken together, these results highlight that the hybrid design offers a practical deployment strategy: when feature extraction can be performed offline or amortized across batches, the lightweight classifiers deliver near-instantaneous predictions, making the approach highly suitable for real-time or resource-constrained environments (e.g., edge devices). Meanwhile, the end-to-end design provides the benefit of a unified pipeline, but at the expense of higher per-sample inference latency. Although the absolute differences may be only a few milliseconds, such gains still represent a meaningful advantage in scenarios requiring rapid or repeated inference.

\subsection{Comparison with previous studies}

\begin{figure}[H]
    \centering
    \includegraphics[width=0.8\textwidth]{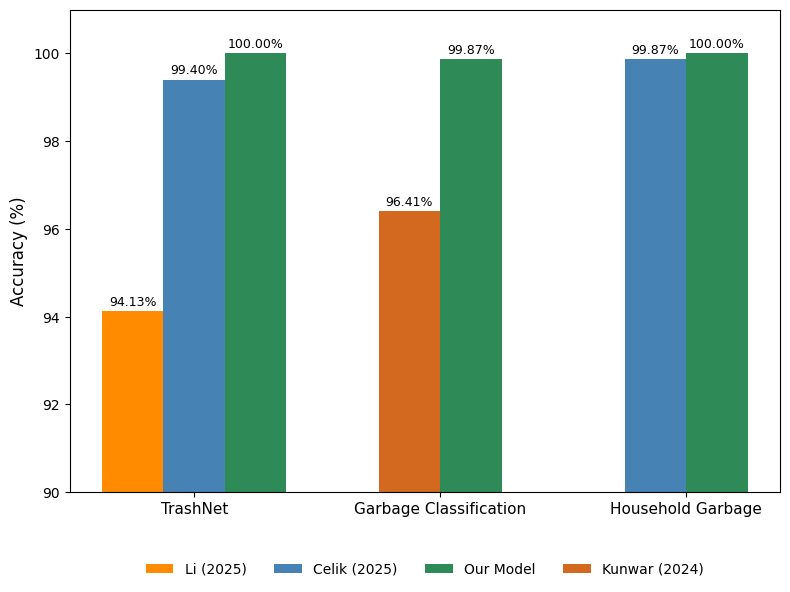}
    \caption{Comparison of our proposed hybrid model with previous methods (SOTA) across three datasets: TrashNet, Garbage Classification, and Household Garbage}

    \label{fig:benchmark_chart}
\end{figure}

In addition, we conducted a comparative analysis against prior works to highlight the advantages of our proposed approach. As summarized in Figure~\ref{fig:benchmark_chart}, the hybrid models consistently outperform existing benchmarks across three datasets.

On the large-scale Garbage Classification dataset, our best hybrid configuration (customized ResNet50 + LoR/SVM) achieves a new state-of-the-art accuracy of 99.87\%, clearly surpassing the 96.41\% reported by ~\citep{kunwar2024} using EfficientNetV2S with hyperparameter optimization.

The generalization ability of our model is further confirmed on the TrashNet dataset. ~\citep{celik} reported 99.40\% using a multi-feature ensemble (EfficientNetB0 + InceptionV3 + HyperColumn). In contrast, our refined hybrid model (customized ResNet50 + LoR) attains 100\% accuracy. Meanwhile, ~\citep{li2025} proposed an improved ResNet50 with redundancy-weighted feature fusion and depthwise separable convolutions, but their method reached only 94.13\% on TrashNet, significantly lower than both Celik's and our hybrid results.

Overall, these results demonstrate that our hybrid framework consistently provides robust and reliable improvements across datasets of different scales and characteristics.

\section{Discussion}
\label{sec:discussion}
The results demonstrate the effectiveness of the proposed hybrid framework while emphasizing both accuracy and efficiency. By combining deep features extracted from a customized ResNet50 backbone with classical machine learning classifiers, the hybrid approach consistently improved performance across all three public datasets. This synergy arises from the complementary strengths of the two components: the fine-tuned ResNet50 acts as a robust feature extractor, whereas lightweight classifiers such as SVM and LoR establish compact and reliable decision boundaries within the learned feature space. Notably, these classifiers are less sensitive to label noise, as confirmed on the corrected Household Garbage dataset, where the hybrid model maintained 100\% accuracy while end-to-end deep learning models experienced minor degradation. This outcome highlights both the robustness of the hybrid paradigm and the importance of dataset integrity for reliable benchmarking.

Feature-selection experiments further revealed that over 95\% of extracted features are redundant. A compact subset of 50–100 features preserved or slightly improved accuracy compared with the full 2048-dimensional representation, reducing computational cost without compromising performance. While CNN-based feature extraction remains the dominant inference cost (6–10\,ms per image), the subsequent classification stage contributes only a negligible overhead (approximately 0.002\,ms per image). This efficiency, combined with near-perfect accuracy, underscores the potential of the hybrid framework for deployment on resource-constrained platforms such as real-time waste classification systems.

Despite these promising results, several limitations remain. The label correction applied to the Household Garbage dataset was limited in scope, and a more exhaustive review would strengthen benchmark reliability. Moreover, the datasets used in this study consist of clean, well-lit, and isolated objects that do not fully reflect real-world waste streams where items may overlap, deform, or appear under poor illumination. Finally, the current pipeline focuses solely on classification of pre-cropped images, whereas practical applications would require integration with object detection and segmentation modules. Addressing these aspects represents a promising direction for future research aimed at enhancing the robustness and applicability of the proposed approach.

\section{Conclusion}
\label{sec:conclusion}
This study investigated automated waste classification by systematically comparing three paradigms: classical machine learning, deep learning, and a hybrid framework. The results demonstrate that the hybrid approach, which leverages customized ResNet50 as a feature extractor, consistently achieved the highest performance across all datasets, reaching 99.87\% accuracy on the Garbage Classification dataset and perfect 100\% accuracy on both TrashNet and the corrected Household Garbage dataset. In addition to accuracy, this study emphasizes the importance of data integrity and computational efficiency. Our analysis shows that top-level performance can be obtained without high computational cost, as feature selection enabled more than 95\% dimensionality reduction while preserving accuracy, resulting in substantially faster training and inference. This combination of accuracy, robustness to label noise, and efficiency highlights the practical potential of the proposed framework for real-world deployment. Future work will focus on incorporating explainable AI techniques for model interpretation, integrating deep and handcrafted features, and extending the proposed framework to diverse domains beyond waste classification.

\section*{Acknowledgments}
This research received no funding.

\section*{Declaration of competing interest}
The authors declare that they have no known competing financial interests or personal relationships that could have appeared to influence the work reported in this paper.

\section*{Declaration of Generative AI and AI-assisted Technologies in the Writing Process}
During the preparation of this manuscript, the authors used AI-assisted tools, including ChatGPT, Gemini, and Grammarly, to improve grammar, refine phrasing, and enhance overall clarity. Following the use of these tools, all authors carefully reviewed and revised the text to ensure accuracy and coherence. The authors take full responsibility for the content of this publication.

\section*{CRediT authorship contribution statement}
\textbf{Ngoc-Bao-Quang Nguyen}: Conceptualization, Methodology, Software, Validation, Data curation, Formal analysis, Writing - original draft, Writing - review \& editing.
\textbf{Tuan-Minh Do}: Conceptualization, Methodology, Visualization, Software, Formal analysis, Writing - original draft, Writing – review \& editing.
\textbf{Cong-Tam Phan}: Software, Validation, Formal analysis, Visualization, Writing - original draft, Writing - review \& editing.
\textbf{Thi-Thu-Hong Phan}: Supervision, Conceptualization, Methodology, Validation, Formal analysis, Project administration, Writing - review \& editing.

\section*{Data availability}
All datasets used in this study are publicly available. The datasets can be accessed via the following links:

\begin{itemize}
    \item \textbf{Garbage Classification dataset:} \url{https://www.kaggle.com/datasets/sumn2u/garbage-classification-v2}
    \item \textbf{Household Garbage dataset:} \url{https://www.kaggle.com/datasets/mostafaabla/garbage-classification}
    \item \textbf{TrashNet dataset:} \url{https://www.kaggle.com/datasets/feyzazkefe/trashnet}
\end{itemize}

\bibliographystyle{elsarticle-num}
\bibliography{ref.bib}

\end{document}